\documentclass[letterpaper, 10 pt, conference]{ieeeconf}
\IEEEoverridecommandlockouts                              
\let\labelindent\relax
\usepackage{graphicx}
\usepackage{amsmath}
\usepackage{amssymb}
\usepackage{enumerate}
\usepackage{enumitem}
\usepackage{algorithm}
\usepackage{algorithmic}
\usepackage{multirow}
\usepackage{xcolor}
\usepackage{color}

\usepackage{colortbl}
\usepackage{dsfont}
\usepackage{framed}
\usepackage{diagbox}
\usepackage{float}
\usepackage{soul}
\usepackage{bbding}
\usepackage{soul}
\usepackage{caption}
\usepackage{graphicx}
\usepackage{diagbox}

\begin{document}
	\title{
		Critical Example Mining for Vehicle Trajectory Prediction using Flow-based Generative Models
	}
	\author{
		Zhezhang~Ding
		and Huijing~Zhao%
		\thanks{This work was supported in part by the National Natural Science Foundation of China under Grant 61973004 and U22A2061. Correspondence: H. Zhao, zhaohj@cis.pku.edu.cn.}
		\thanks{
			Z. Ding and H. Zhao are with the Key Lab of Machine Perception (MOE) and also with the School of Intelligence Science and Technology, Peking University, Beijing, China. (e-mail: zzding@pku.edu.cn; zhaohj@pku.edu.cn).}
	}
	\maketitle
	\thispagestyle{empty}
	\pagestyle{empty}
	
	\begin{abstract}
		\bfseries Precise trajectory prediction in complex driving scenarios is essential for autonomous vehicles. In practice, different driving scenarios present varying levels of difficulty for trajectory prediction models. However, most existing research focuses on the average precision of prediction results, while ignoring the underlying distribution of the input scenarios. This paper proposes a critical example mining method that utilizes a data-driven approach to estimate the rareness of the trajectories. By combining the rareness estimation of observations with whole trajectories, the proposed method effectively identifies a subset of data that is relatively hard to predict BEFORE feeding them to a specific prediction model. The experimental results show that the mined subset has higher prediction error when applied to different downstream prediction models, which reaches +108.1\% error (greater than two times compared to the average on dataset) when mining 5\% samples. Further analysis indicates that the mined critical examples include uncommon cases such as sudden brake and cancelled lane-change, which helps to better understand and improve the performance of prediction models.

	\end{abstract}
	\section{Introduction}
	The past decades have witnessed rapid and tremendous developments in autonomous vehicles \cite{ma2020artificial}. To achieve safe and effective autonomous driving in real traffic, the autonomous vehicles should understand and predict the possible motions of all factors in the scene, which are categorized as Trajectory Prediction (TP) or Vehicle Trajectory Prediction (VTP) in related research \cite{ding2023incorporating}. 
	
	A lot of models and methods have been developed to solve the TP problem \cite{mozaffari2020deep}. 
	Earlier methods exploit probabilistic \cite{Brechtel2014} or game-theory methods \cite{Bahram2016} to model the movement of all factors in the scene, while recent methods use deep neural network techniques, such as RNN \cite{zyner2019naturalistic}, GNN \cite{gu2021densetnt}, or Transformers \cite{ngiam2022scene}, to directly predict the future trajectory in end-to-end manner. 
	The performance of these methods is mostly evaluated on public datasets using metrics such as RMSE (Root Mean Square Error) or ADE (Average Displacement Error), and improving overall performance such as average prediction accuracy on the dataset has been the focus of the majority of literature works.
	\begin{figure}[tb]
		\begin{center}
			\includegraphics[keepaspectratio=true,width=0.75\linewidth]{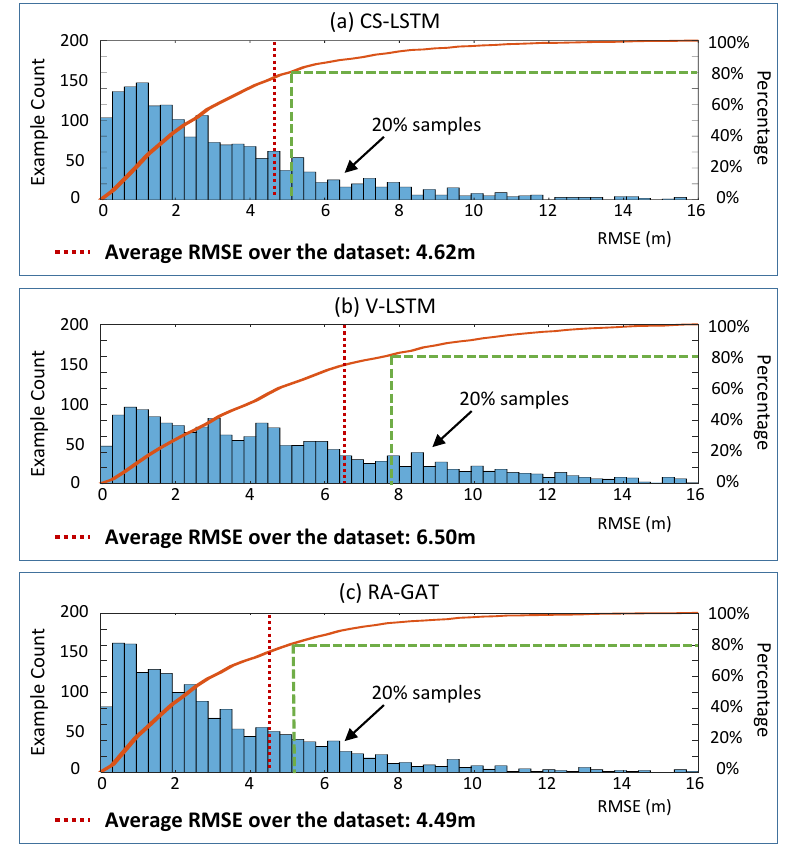}
		\end{center}
		\caption{Long-tail phenomenon in different trajectory prediction models. Nearly 20\% of samples hold extremely higher errors than average. The CS-LSTM and V-LSTM are taken from \cite{deo2018convolutional}, while RA-GAT is introduced in \cite{ding2021ra}.}
		\label{Fig_long}
	\end{figure}

	While the overall performance of VTP models can be fair, all models suffer from high prediction errors on a small portion of the dataset, namely \textbf{Critical Examples}. Fig.\ref{Fig_long} shows the histograms of prediction errors for several typical trajectory prediction models on a subset of the NGSIM dataset, which are very long-tailed: with 80\% of the examples having high prediction accuracies, while the errors of the rest 20\% are significantly higher than the average of the dataset. The average prediction accuracy on datasets cannot conclusively indicate whether the model effectively addresses critical examples, or if it simply performs marginally better on typical cases.	
	
	The problem has been referred to as long-tail learning \cite{zhang2023deep} or learning on imbalanced dataset \cite{he2009learning}, which has been well studied in literature for classification tasks. These works address the problem where the frequent and tail classes are distinct and clearly defined. For example, for 3D detection task, \cite{jiang2022improving} provides a solid investigation of rareness definition and estimation, where referring the rare instances such as irregularly shaped vehicles or pedestrians as the intra-class long-tail, a rareness measure is developed via a flow-based generative model. However, in the scope of VTP, how to identify and model critical examples is still an open question. There exists some researches that address similar problems targeting VTP task. \cite{zhou2022long} builds an uncertainty-based framework to mitigate the problem of sparse or unavailable training data that leads to downstream planner failures. \cite{makansi2021exposing} evaluates the difficulty of samples by the prediction error of a Kalman filter, and proposes a supplementary loss that places challenging cases closer in the feature space, \cite{kozerawski2022taming} similarly designs two moment-based tailedness measurement for prediction.

While the above works focus on improving prediction performance on long-tail data, this research investigates critical example mining for VTP task. Different from the literature works that use the results of prediction models such as uncertainty or error in criticality assessment, which is \textbf{model-relevant}, this research proposes a \textbf{model-irrelevant} mining criterion by modeling the density of examples so that the mined critical results are not bound up with downstream VTP models. Critical examples are mined in two dimensions: (1) \textbf{Rare Examples}: from the data distribution perspective; (2) \textbf{Hard Examples}: from the VTP task perspective. Inspired by \cite{jiang2022improving}, this research exploits normalizing flows to model the probability density of data, based on which, the rareness and its hardness for VTP task are estimated. To this end, we further design hand-crafted features by driving knowledge to convert raw trajectory data to low-dimensional but efficient input for model learning.
	
We claim our work to have following contributions:
	\begin{enumerate}
		\item We formulate the problem of critical example mining for VTP task from both rareness and hardness dimensions, leading to \textbf{model-irrelevant} mining criteria and results - The experimental results show that the mined examples are relatively critical to all the downstream prediction models.  
		\item We use normalizing flows to model probability density of VTP data, based on which the criteria of rareness of trajectory data and its hardness for VTP task are estimated. To the best of our knowledge, this is the first work that builds flow-based generative models on VTP data to respectively model the rareness and hardness of the input.
		\item We design hand-crafted features by making use of driving knowledge and convert raw VTP data to low-dimensional but efficient input of flow-based model, extensive experiments are conducted to demonstrate that the method is capable to dig out samples with +108.1\% error (greater than two times) compared to the average on the dataset and find out uncommon cases such as sudden braking and cancelled lane-change.
	\end{enumerate}
	
	The remainder of the paper is organized as follows. Section II gives a literature review, section III describes the details of the proposed approach of critical example mining. Experimental results are presented in section IV, followed by conclusion and future works in section V.
	
	\section{Related Works}
	\label{Sec_relatedwork}
	\subsection{Vehicle Trajectory Predcition}
	Trajectory Prediction is one of the fundamental tasks in the field of autonomous vehicles. It takes the environment observation as input, and the future motions of ego/surrounding vehicles are set to be the output. Specifically, researchers try to consider many driving-related features during the inference of trajectory, such as road structure \cite{gao2020vectornet}, maneuver \cite{deo2018convolutional}, and interaction patterns between vehicles \cite{lee2019joint}, which are categorized as Driving Knowledge in \cite{ding2023incorporating}.
	
	The trajectory prediction models can be roughly divided into modeling-based and Deep-neural-network-based methods. The modeling-based methods consist of earlier works that exploit probabilistic \cite{Brechtel2014} or game-theoretic \cite{Bahram2016} manner to explicitly model the motion of all factors in the scene. The Deep-neural-network-based methods consist of recent works that use different networks to implicitly infer the future trajectory of the target vehicle. The backbone networks in these methods 
	evolve from LSTM \cite{altche2017lstm}, CNN \cite{messaoud2019relational}, to GNN \cite{xu2022group} and Transformers \cite{huang2022multi}.
	
	However, with the networks evolve, most of the existing works mainly focus on the average prediction performance on public dataset, while the undelying distribution variance of trajectory data is ignored.
	\subsection{Critical Example Mining}
	Under the circumstances, it is crucial to dig out the inputs of great importance from the original dataset, which is categorized as Critical Example Mining in this work.
	
	Through the limited discussion on this perspective, one possible solution is to estimate the rareness or the hardness of the input sample. \cite{zhou2022long} describe the proportion of naturalistic driving data as "long-tail" phenomenon, in which the majority of samples describe normal driving cases, while the minority of samples are the genuine critical samples that are hard for the model to predict. The author exploits an uncertainty-based methods to detect the rare samples. \cite{ding2021ra} analyses the different prediction performances between various maneuver samples, the result shows that the sudden brake in car-following or the moment before lateral movement in a lane-change process are the subgroups with relatively higher prediction error. However, none of the above works explicitly distinguish the difference between "rareness" and "hardness", while we believe these two dimensions are quite different and need to be considered separately.
	
	Another possible way to find out the critical inputs is through dataset distillation manner \cite{yu2023dataset}, which is originally designed to derive a much smaller dataset with samples but still have comparable performance when applying the selected samples during training \cite{wang2018dataset}. However, such a process is not available  during test time, and the results bind to the downstream model itself.
		\begin{figure*}[]
		\begin{center}
			\includegraphics[keepaspectratio=true,width=0.95\linewidth]{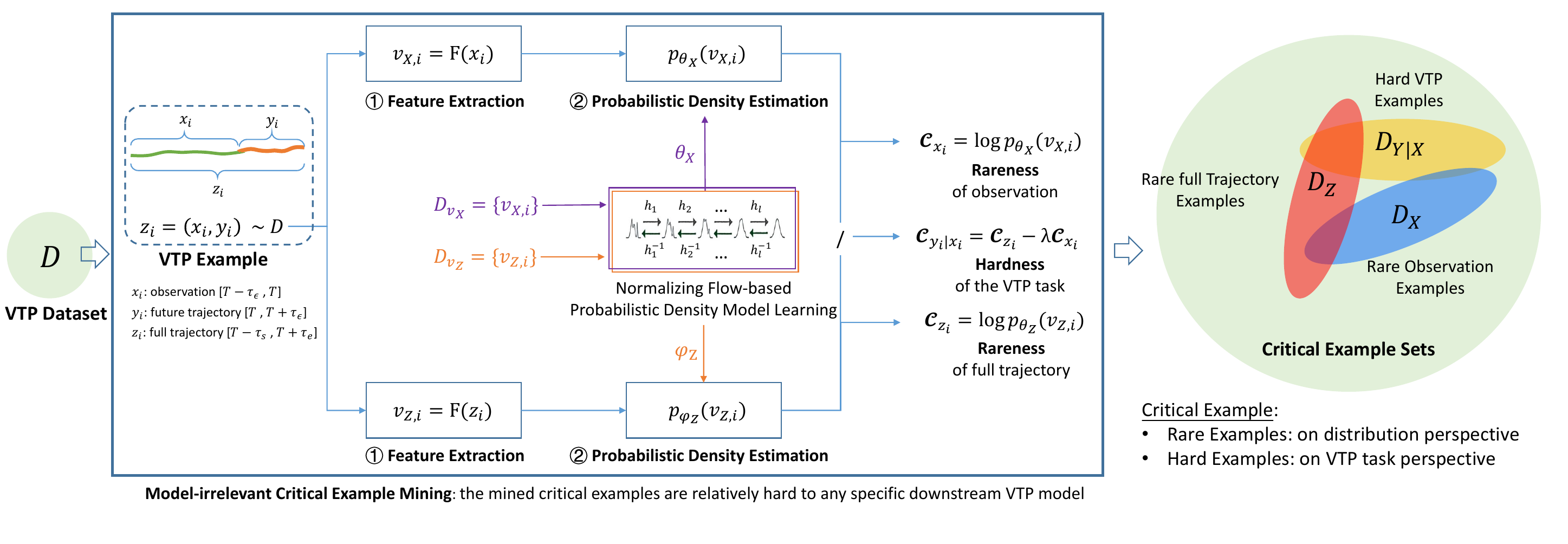}
		\end{center}
		\caption{Outline of the proposed critical example mining method. The details of \textcircled{1} Feature Extraction will be introduced in Sec.\ref{Subsec_feature}), the \textcircled{2} Probability Density Estimation will be expanded in Sec.\ref{Subsec_flow}.}
		\label{Fig_overview}
	\end{figure*}
	
	\subsection{Generative Models}
	
	To achieve critical example mining in a model-irrelevant manner, we have to start with the original input distribution. However, as the ground truth of the distribution is not observable, we exploit generative models to approximate such distribution. The generative models are designed to learn the underlying patterns and distribution of the training dataset, and then use the learned hidden pattern to generate new data \cite{harshvardhan2020comprehensive}.

	There are different kinds of generative models in literature. One of the most popular generative models is the Variational AutoEncoder (VAE), which learns the hidden state from training dataset under guassian assumption \cite{kingma2019introduction}. Another popular variant is Generative Adversarial Networks (GAN), which train a generator and a discriminator iteratively \cite{goodfellow2020generative}. However, if the dimension of the learned feature space is too high, we still cannot easily distinguish which sample is "more critical". 
	In this work, we exploit another variant of generative networks named Flow-based generative model \cite{rezende2015variational}, which can approximate the distribution of the training data via a simple but effective distribution. Under such formulation, it is possible to estimate the criticality in a low-dimension feature space.

	\section{Methodology}
	\label{Sec_method}
	\subsection{Problem Formulation}
	\label{Subsec_formulation}
	The problem of vehicle trajectory prediction (VTP) can be generally formulated in a probabilistic way as estimating a conditional distribution
	\begin{equation}
		\label{eqn_1}
		P(Y|X)
	\end{equation}
	where $X$ is the observation of driving state consisting of both the target vehicle(s) and scene context up to the prediction time $t$, e.g. $X=\{X_{T-\tau_s},...,X_T\}$, $Y$ is the predicted trajectories of the target vehicle(s) in a future time horizon, e.g. $Y=\{Y_{T+1},...,Y_{T+\tau_e}\}$.
	At time $T$ for prediction, $X$ describes the past and the current, whereas $Y$ is the future. Let $Z=(X,Y)$ be the joint $X-Y$ pair, describing the entire driving course, and we have
	\begin{equation}
		\label{eqn_2}
		P(Z)=P(Y|X)P(X)
	\end{equation}
	Given a data set $D = \{z_i\}^N_{i=1}$ and a mining ratio $r$, where $z_i=(x_i,y_i)$ and $x_i,y_i,z_i$ are data samples of variables $X$, $Y$ and $Z$ respectively.
	Our purpose of critical examples mining is threefold, namely to mine sets of
	\begin{itemize}
		\item {\textbf{Rare} examples of observation $X$
			
		}
	\begin{equation}
		D_X^*= \underset{D_X \subset D, ||D_X||=r*||D||} {\arg\min} \mathbf{E}_{z_i \sim D_X}(\boldsymbol{\mathcal{C}_{x_i}})
	\end{equation}

		\item {\textbf{Rare} examples of full trajectory $Z$
			
		}
	\begin{equation}
		D_Z^*= \underset{D_Z \subset D, ||D_Z||=r*||D||} {\arg\min} \mathbf{E}_{z_i \sim D_Z}(\boldsymbol{\mathcal{C}_{z_i}})
	\end{equation}
		\item {\textbf{Hard} examples for trajectory prediction $Y|X$
			
		}
	\end{itemize}
	 \begin{equation}
	D_{Y|X}^*= \underset{D_{Y|X} \subset D, ||D_{Y|X}||=r*||D||} {\arg\min} \mathbf{E}_{z_i \sim D_{Y|X}}(\boldsymbol{\mathcal{C}_{y_i|x_i}})
	\end{equation}
	The $\boldsymbol{\mathcal{C}_{x_i}}$ and $\boldsymbol{\mathcal{C}_{z_i}}$ represent the rareness estimation of $x_i$ and $z_i$ respectively, while $\boldsymbol{\mathcal{C}_{y_i|x_i}}$ represents the hardness of predicting $y_i$ given $x_i$. These indicators are implemented based on the probability density of $X$ and $Z$, which will be introduced in Sec.\ref{Subsec_minginmethod}).

	As the genuine distribution of $X$ and $Z$ is hard to obtain, we exploit normalizing flow to model the probability density $p_X$ and $p_Z$ in this research. To make the learning process more effective and VTP-oriented, we design hand-crafted features by making use of driving knowledge, and build low-dimensional input on extracted features $v_{X,i}=F(x_i), x_i \sim X$ and $v_{Z,i}=F(z_i), z_i \sim Z$ from raw trajectory data, and learn $p_{\theta_X}$ and $p_{\phi_Z}$ on datasets $D_{v_X}=\{v_{X,i}\}^N_{i=1}$ and $D_{v_Z}=\{v_{Z,i}\}^N_{i=1}$ respectively. Below, we describe the methods of normalizing flow-based probability density model learning in Sec.\ref{Subsec_flow}, feature extraction and mining method in Sec.\ref{Subsec_feature}).

	\subsection{Probability Density Modeling using Normalizing Flow}
	\label{Subsec_flow}
	Given a feature vector $v$ of an example, we use normalizing flow to estimate the log probability density $\log p_{\theta}(v)$ as below.
	Under the definition of normalizing flow, we use a learned invertible function $h$ to transform the input $v$ into a latent variable $b$ that follows a base distribution $p(b)$. Thus, we have
	
	\begin{equation}
		b = h(v)
	\end{equation}
	and the inverse process of $h$ can be regarded as sampling or generating $v$ from the latent $b$, such that
	\begin{equation}
		v = h^{-1}(b)
	\end{equation}
	
	The base distribution $p(b)$ is generally chosen to be an analytically tractable distribution whose probability density function can be easily computed. 	
	Under the change-of-variable rule, the distribution of $v$ can be calculated as:
	\begin{equation}
		\label{flow_fund}
		p_{\theta}(v) = p(b) |det \frac{\partial h(v)}{\partial v}|
	\end{equation}
	where $\theta$ concludes all of the learnable parameters in the whole flow-based model.
	
	In practice, the overall transformation $h$ is composed of several connected invertible layers, such that:
	\begin{equation}
		b = h(v) = h_1 \circ h_2 \circ ... \circ h_l(v)
	\end{equation}
	where $l$ represents the number of layers in the model.
	
	Under such definition, formula (\ref{flow_fund}) can be extended to
	\begin{equation}
		p_{\theta}(v) = p(b) \prod_{i=1}^l |det \frac{\partial h_i(v)}{\partial v}|
	\end{equation}
	
	The log-likelihood of $v$ can be then estimated as:
	\begin{equation}
		\label{flow_final}
		\log p_{\theta}(v) = \log p(b) + \sum_{i=1}^{l}\log|det \frac{\partial h_i(v)}{\partial v}|
	\end{equation}
	
	Formula (\ref{flow_final}) is computable when both $p(b)$ and $|det \frac{\partial h_i(v)}{\partial x}|$ are easy to calculate. For implementation, we exploit NICE model \cite{dinh2014nice} in experiment.
	
	Given the extracted sets $D_{v_X}=\{v_{X,i}\}^N_{i=1}$ and $D_{v_Z}=\{v_{Z,i}\}^N_{i=1}$ of feature vectors, learning the probability density models $p_{\theta_X}$ and $p_{\phi_Z}$ can be formulated as minimizing the expected negative log-likelihood as below:
	
	\begin{eqnarray}
		\theta_X= \underset{\theta} {\arg\min} \mathbf{E}_{v_i \sim D_{v_X}}[-\log p_{\theta}(v_i)] \\
		\phi_Z= \underset{\phi} {\arg\min} \mathbf{E}_{v_i \sim D_{v_Z}}[-\log p_{\phi}(v_i)]
	\end{eqnarray}

		\subsection{Feature Extraction and Mining Method}
	
	\subsubsection{Feature Extraction}
	\label{Subsec_feature}
\begin{figure}[]
		\begin{center}
			\includegraphics[keepaspectratio=true,width=0.8\linewidth]{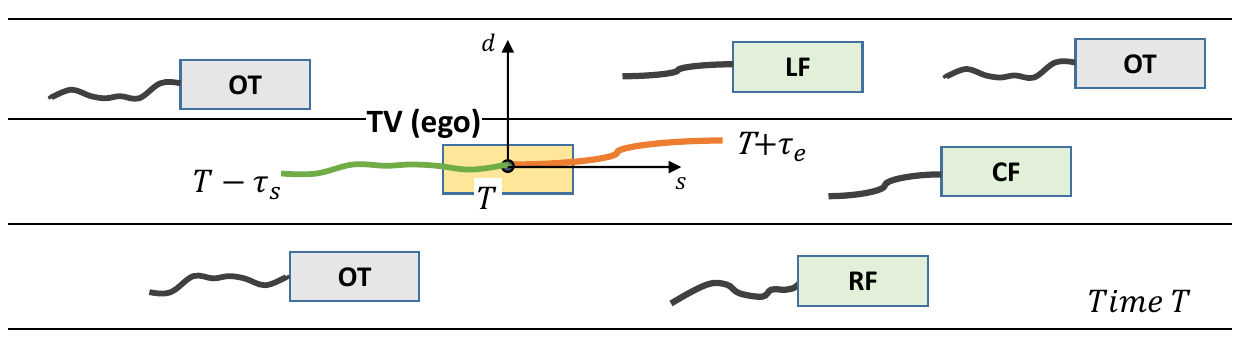}
		\end{center}
		\caption{Relevant scene vehicles for TV's feature extraction.}
		\label{Fig_feat}
	\end{figure}

	Consider the task of predicting the future trajectory of a target vehicle (TV) at time $T$. 
	In addition to the driving state of TV up to the current frame, scene context has a non-negligible impact on its future driving trajectory. In on-road driving scenario, the most influential scene factors are the left front (LF), the center front (CF), and the right front (RF) vehicles of the TV. Therefore, a driving scene is factorized using the features extracted from the trajectory data of the TV, LF, CF, and RF, as illustrated in Fig.\ref{Fig_feat}.

	In this research, we design a uniform feature extraction method and feature vector format for $v_{X,i}=F(x_i), x_i \sim X$ and $v_{Z,i}=F(z_i), z_i \sim Z$, so that the probability density model $p_{\theta_X}$ and $p_{\phi_Z}$ can be learned using the same method with comparable scale. 
	Each data example $d$ is regularly partitioned into $M$ segments over its time span. In case $d \in D_X$, the time interval is $[T-\tau_s,T]$, and the length (in second) is $T_s=\tau_s$. In case $d \in D_Z$, the time interval is $[T-\tau_s,T+\tau_e]$, and the length (in second) is $T_s=\tau_s+\tau_e$. The following two partition methods are exploited and compared in experiments.
	\begin{itemize}
		\item Fixing the number of segments, FixSegNum ($\kappa$), i.e. example $d$ is partitioned into $\kappa$ segments, and the length of each segment is $T_s/\kappa$;
		\item Fixing the length (in seconds) of segments, FixSegLen ($\kappa$), i.e. example $d$ is partitioned into $T_s/\kappa$ segments, and the length of each segment is $\kappa$.
	\end{itemize}
	
	From each segment, the following descriptive features are extracted. 
	\begin{itemize}
		\item  Velocity ($f_1-f_{10}$): Velocity of $TV$, $LF$, $CF$, $RF$, and the average of other scene vehicles $OT$. Each velocity is a two-dimensional vector containing lateral and longitudinal components;
		\item  Acceleration ($f_{11}-f_{20}$): Acceleration of the $TV$, $LF$, $CF$, $RF$, and the average acceleration of $OT$. Each acceleration is a two-dimensional vector containing lateral and longitudinal components;
		\item  Gap Distance ($f_{21}-f_{23}$): Gap distance from $TV$ to $LF$, $CF$, $RF$. In this research, only the gap on longitudinal direction is considered;
		\item  Relative Velocity ($f_{24}-f_{26}$): Relative velocity of $TV$ compared to $LF$, $CF$, $RF$. In this research, only the longitudinal component is considered.
	\end{itemize}
	If one of the scene factors does not exist, is not available for feature estimation, or is not adapted to the example data, the feature value in the corresponding dimension will be assigned a fixed value. 
	By sequentially aligning $f_1-f_{26}$ of all $M$ segments, a $26*M$-dimensional feature vector $v$ is extracted.
	
	\subsubsection{Mining Method}
	\label{Subsec_minginmethod}
	Combing the processes of probability density modeling and feature extraction, we finally calculate
	\begin{equation}
	\boldsymbol{\mathcal{C}_{x_i}} = \log p_{\theta_X}(v_{X,i}),
	\boldsymbol{\mathcal{C}_{z_i}} = \log p_{\phi_Z}(v_{Z,i})
	\end{equation}
	The lower value indicates lower probability density, i.e. higher rareness.
	
	As for hardness, we have a heuristic hypothesis that when the observations are normal (high $\boldsymbol{\mathcal{C}_{x_i}}$) while the whole trajectory is rare (low $\boldsymbol{\mathcal{C}_{z_i}}$), such a case is relatively hard to predict.
	
	Thus, the estimation of $\boldsymbol{\mathcal{C}_{y_i|x_i}}$ can be defined as 
	\begin{equation}
	\label{lambda}
	\boldsymbol{\mathcal{C}_{y_i|x_i}} = \boldsymbol{\mathcal{C}_{z_i}} - \lambda\boldsymbol{\mathcal{C}_{x_i}}
	\end{equation}
	where $\lambda$ is an adjustable scaling parameter. The lower value of $\boldsymbol{\mathcal{C}_{y_i|x_i}}$ describes higher hardness.
	
	Following the above settings, the optimization process to obtain $D_X^*$, $D_Z^*$ and $D_{Y|X}^*$ in formula (3)-(5) are implemented by ranking samples based on $\boldsymbol{\mathcal{C}_{x_i}}$, $\boldsymbol{\mathcal{C}_{z_i}}$, and $\boldsymbol{\mathcal{C}_{y_i|x_i}}$ respectively, then collecting the top $r$ percent samples with minimum values, i.e.
	
	\begin{equation}
	\label{threshold}
	\begin{split}
		D_X^*  &=  \{z_i | \boldsymbol{\mathcal{C}_{x_i}}< \delta_X \} \\
		D_Z^* & =  \{z_i | \boldsymbol{\mathcal{C}_{z_i}}< \delta_Z \} \\
		D_{Y|X}^* & = \{z_i | \boldsymbol{\mathcal{C}_{y_i|x_i}}< \delta_Y \} 
	\end{split}
	\end{equation}
	where $\delta_X$,$\delta_Z$,$\delta_Y$ represents the $r$-percentile of $\boldsymbol{\mathcal{C}_{x_i}}$, $\boldsymbol{\mathcal{C}_{z_i}}$ and $\boldsymbol{\mathcal{C}_{y_i|x_i}}$ respectively.
	\section{Experiment}
	\label{Sec_Exp}
	\subsection{Dataset}
	The trajectory prediction task is widely tested on two public datasets: NGSIM I-80\cite{I80} and US-101\cite{US101}. Both datasets contain real freeway driving data under mild, moderate, and congested traffic flow. The original data is captured at 10 Hz from top-down view via surveillance cameras. We conduct our mining method based on such trajectories following the setting of 8 seconds length for $Z$, with 3s as input history $X$ and 5s for prediction ground truth $Y$. We then randomly sample 2000 trajectories for model learning and evaluation, denoted as $NGSIM_{2K}$.  
	
	\subsection{Experimental Settings}
	\label{Subsec_metric}
	\subsubsection{Reference Label Definition}
	Before validating and comparing the mining results, it is necessary to establish a definition for the term 'critical sample' in a model-irrelevant manner. As manual labelling of criticality is not feasible, we adopt the approach used in \cite{makansi2021exposing} to utilize the constant velocity Kalman Filter to generate reference labels for the samples. We use symbol $D^{ref}_r$ to include the top $r$ percent samples with highest error under the Kalman Filter in $NGSIM_{2K}$.
	
	\subsubsection{Downstream Trajectory Prediction Models}
	To compare the performance of mining, we apply the mined subset on different downstream prediction models then take the prediction error as metrics to examine the effect of mining. The downstream models included in this paper are:
	\begin{itemize}
		\item CS-LSTM: The convolutional social pooling LSTM introduced in \cite{deo2018convolutional}. Abbreviated as $CS$.
		\item Vallina-LSTM: The simple LSTM prediction model in \cite{deo2018convolutional}. Abbreviated as $V$.
		\item RA-GAT: The repulsion and attraction graph attention model introduced in \cite{ding2021ra}. Abbreviated as $RA$.
	\end{itemize}
	We use symbol $D^{CS}_r$ to include the top $r$ percent of samples with highest prediction error when applying CS-LSTM to the $NGSIM_{2K}$ dataset. $D^{V}_r$ and $D^{RA}_r$ are also defined similarly for Vallina-LSTM and RA-GAT. 
	
	\subsubsection{Mining Baselines}
	In Sec.\ref{Subsec_formulation}, we define 3 different mining subset $D_X^*$, $D_Z^*$ and $D_{Y|X}^*$. 
	We compare the three mining results with two different baselines:
	\begin{itemize}
		\item $D^{rand}_r$: Randomly select $r$ percent samples from the original dataset. 
		\item $D^{ref}_r$: Collecting top $r$ samples with highest error through Constant Velocity Kalman Filter. Note that we take $D^{ref}_r$ both for reference labels and for model-irrelevant mining baseline.
 	\end{itemize} 
 
 	\subsubsection{Metrics}
	The following factors are used to estimate the effect of a given mined result $\hat{D}_r$ from original $D$, where $\hat{D}\in\{D_X^*, D_Z^*,D_{Y|X}^*,D^{rand}_r,D^{ref}_r\}$ and subscript $r$ denotes the mining ratio.
	\begin{itemize}
		\item $Err(\hat{D})^{M}$: 5 seconds RMSE prediction error when applying $\hat{D}_r$ on downstream model $M$, $M \in\{CS,V,RA\}$. $Err(D)^{M}$ represents the 5 seconds error on the original dataset $D$.
		
		\item $\triangle Err$: the percentage of 5s prediction error change when applying $\hat{D}_r$ to model $M$ compared to $D$, computed as 
		\begin{equation}
			\triangle Err(M) = \frac{Err(\hat{D}_r)^{M} -  Err(D)^{M}}{Err(D)^{M}},
		\end{equation}
		 $M\in \{CS,RA,V\}$ denotes the downstream model. 
		
		\item $Cov$ : \textbf{Model-relevant} mining coverage of $D^{M}_r$, calculated as:
		\begin{equation}
			Cov(M) = \frac{||\hat{D}_r \cap D^{M}_r||}{||D^{M}_r||}
		\end{equation}
		$M\in \{CS,RA,V\}$ denotes the downstream model.
		\item $Cov^{ref}$ : \textbf{Model-irrelevant} mining coverage of the reference label $D^{ref}_r$, computed as 
		\begin{equation}
			Cov^{ref} = \frac{||\hat{D}_r \cap D^{ref}_r||}{||D^{ref}_r||}
		\end{equation}
	\end{itemize}  

\begin{figure*}[tb]
	\begin{center}
		\includegraphics[keepaspectratio=true,width=0.7\linewidth]{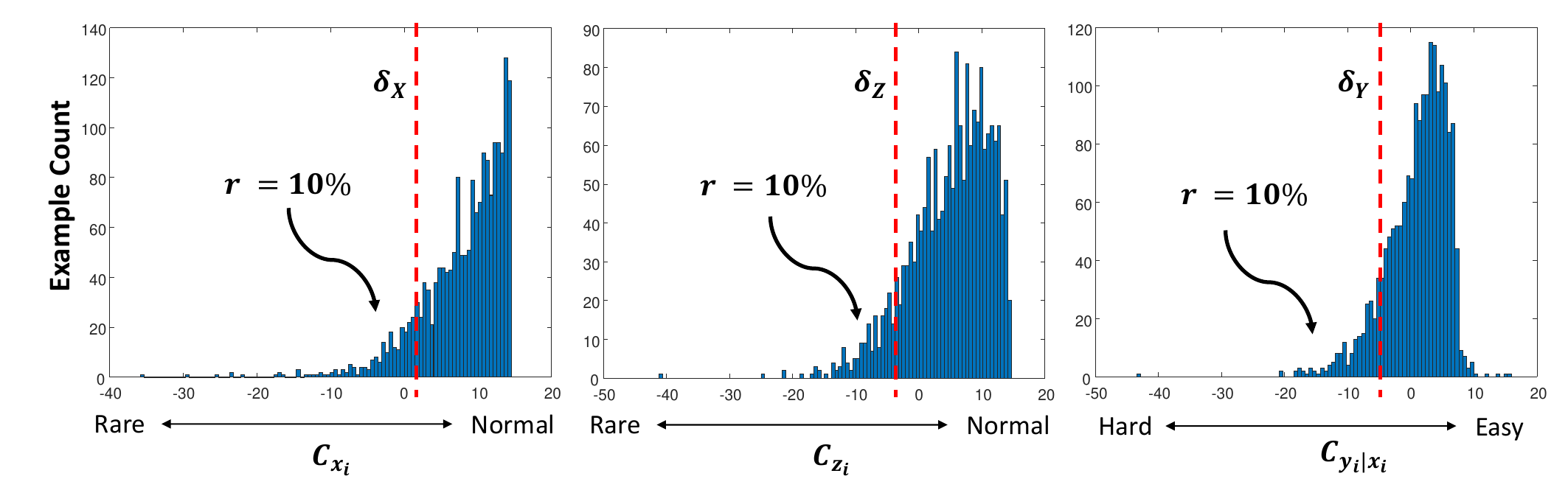}
	\end{center}
	\caption{Learning results of $\boldsymbol{\mathcal{C}_{x_i}}$ , $\boldsymbol{\mathcal{C}_{z_i}}$, and $\boldsymbol{\mathcal{C}_{y_i|x_i}}$. The threshold $\delta_X$, $\delta_Z$, and $\delta_Y$ with $r=10\%$ are showed by the red dashed-lines.}
	\label{Fig_score}
\end{figure*}
	\begin{table*}[]
	
	\caption{Prediction Error and Error Change of Mined Subset on $RA-GAT$}
	\centering
	\label{Tab_mining}
	\resizebox{0.7\linewidth}{!}
	{
		\begin{tabular}{|c|cc|cc|cc|cc|cc|c|}
			\hline
			\multirow{2}{*}{\textbf{r}} & \multicolumn{2}{c|}{\textbf{$D_X^*$}}                & \multicolumn{2}{c|}{\textbf{$D_Z^*$}}                & \multicolumn{2}{c|}{\textbf{$D_{Y|X}^*$}}                & \multicolumn{2}{c|}{\textbf{$D_{rand}^r$}}            & \multicolumn{2}{c|}{\textbf{$D_{ref}^*$}}                &  \multirow{2}{*}{\begin{tabular}[c]{@{}c@{}}$Err(D)$)\\ (r = 100\%)\end{tabular}} \\ \cline{2-11} 
			& \multicolumn{1}{c|}{$Err$}   & $\triangle Err$ & \multicolumn{1}{c|}{$Err$}   & $\triangle Err$ & \multicolumn{1}{c|}{$Err$}   & $\triangle Err$ & \multicolumn{1}{c|}{$Err$}   & $\triangle Err$ & \multicolumn{1}{c|}{$Err$}  & $\triangle Err$ &                              \\ \hline
			20\%               & \multicolumn{1}{c|}{5.398} & +20.1\%     & \multicolumn{1}{c|}{6.620} & +47.2\%     & \multicolumn{1}{c|}{\textbf{6.697}} & \textbf{+49.0\%}     & \multicolumn{1}{c|}{4.410} & -1.90\%    & \multicolumn{1}{c|}{6.679} & +48.5\%         & \multirow{4}{*}{4.496}       \\ \cline{1-11}
			15\%               & \multicolumn{1}{c|}{5.643} & +25.5\%     & \multicolumn{1}{c|}{7.165} & +59.4\%     & \multicolumn{1}{c|}{\textbf{7.333}} & \textbf{+63.1\%}     & \multicolumn{1}{c|}{4.318} & -4.0\%    & \multicolumn{1}{c|}{7.011} & +55.9\%         &                              \\ \cline{1-11}
			10\%               & \multicolumn{1}{c|}{5.470} & +21.7\%     & \multicolumn{1}{c|}{7.919} & +76.1\%     & \multicolumn{1}{c|}{\textbf{8.017}} & \textbf{+78.3\%}     & \multicolumn{1}{c|}{4.301} & -4.3\%    & \multicolumn{1}{c|}{7.834} & +74.2\%          &                              \\ \cline{1-11}
			5\%                & \multicolumn{1}{c|}{6.500} & +44.6\%     & \multicolumn{1}{c|}{9.233} & +105.3\%     & \multicolumn{1}{c|}{\textbf{9.356}} & \textbf{+108.1\%}     & \multicolumn{1}{c|}{4.303} & -4.3\%    & \multicolumn{1}{c|}{8.934} & +98.7\%          &                              \\ \hline
		\end{tabular}
	}
	
\end{table*}
	\subsubsection{Experimental Design}
	The experiments are conducted to answer the following questions.
	\begin{itemize}
		\item How efficient is the normalizing flow-based probability density modeling? 
		\item Are the mined examples critical? Are the criticality relevant or irrelevant to the downstream VTP models?
	\end{itemize}
	In Sec.\ref{Subsec_feature} we give two different feature partition methods, an ablation study is conducted to answer the following question:
	\begin{itemize}
		\item How sensitive is the mining method to feature partition and hyper-parameter setting?
	\end{itemize}
	Finally, we conduct case studies to intuitively demonstrate the examples we mined.
	
	
	
	
	\subsection{Model Learning Results}

	\begin{table*}[]
		\caption{$\triangle Err$ and $Cov$ on Different Downstream Models.}
		\centering
		\label{Tab_modelfree}
		\resizebox{0.6\linewidth}{!}
		{
			\begin{tabular}{|c|c|c|c|c|c|c|c|}
				\hline
				\multicolumn{8}{|c|}{\textbf{$D_X^*$} mining, $FixSegNum(\kappa = 5)$} \\ \hline
				r    & $\triangle Err(RA)$ & $Cov(RA)$ & $\triangle Err(V)$ & $Cov(V)$     &$\triangle Err(CS)$        &  $Cov(CS)$       & $Cov^{ref}$ \\ \hline
				20\%  & +20.1\% & 31.8\%  & +7.4\% & 24.8\%    & +22.2\%   & 30.3\%     & 23.0\%     \\ \hline
				15\% & +25.5\% & 26.3\%  & +9.4\% & 17.3\%  & +27.4\% & 23.3\%    & 18.7\%    \\ \hline
				10\% & +21.7\% & 16.5\%  & +12.0\% & 11.0\% & +25.6\%   & 17.5\% & 13.5\%    \\ \hline
				5\% & +44.6\% & 12.0\%  & +25.6\% & 11.0\%  & +45.5\% & 12.0\%    & 8.0\%    \\ \hline

				\multicolumn{8}{|c|}{\textbf{$D_Z^*$} mining, $FixSegNum(\kappa = 5)$} \\ \hline
				r    & $\triangle Err(RA)$ & $Cov(RA)$ & $\triangle Err(V)$ & $Cov(V)$     &$\triangle Err(CS)$        &  $Cov(CS)$       & $Cov^{ref}$ \\ \hline
				20\%  & +47.2\% & 40.3\%  & +44.6\% & 44.3\%   & +47.9\%    & 38.8\%    & 46.0\%    \\ \hline
				15\% & +59.4\%  & 36.7\%  & +54.2\% & 40.3\%   & +60.4\%    & 36.7\%  & 43.7\%    \\ \hline
				10\% & +76.1\%  & 30.0\%  & +60.1\% & 34.5\% & +75.1\% & 33.0\% & 40.5\%    \\ \hline
				5\% & +105.3\%  & 25.0\%  & +77.9\% & 25.0\%   & +99.9\% & 23.0\% & 34.0\%    \\ \hline
				
				\multicolumn{8}{|c|}{\textbf{$D_{Y|X}^*$} mining, $FixSegNum(\kappa = 5)$} \\ \hline
				r    & $\triangle Err(RA)$ & $Cov(RA)$ & $\triangle Err(V)$ & $Cov(V)$     &$\triangle Err(CS)$        &  $Cov(CS)$       & $Cov^{ref}$ \\ \hline
				20\%  & +49.0\% & 40.5\%  & +53.3\% & 53.5\%   & +50.1\%    & 39.8\%    & 59.0\%    \\ \hline
				15\% & +63.1\%  & 38.7\%  & +65.9\% & 50.3\%   & +63.3\%    & 39.7\%  & 55.7\%    \\ \hline
				10\% & +78.3\%  & 32.0\%  & +77.9\% & 48.0\% & +78.6\% & 36.0\% & 49.5\%    \\ \hline
				5\% & +108.1\%  & 29.0\%  & +102.5\% & 39.0\%   & +106.7\% & 28.0\% & 41.0\%    \\ \hline

			\end{tabular}
		}
	\end{table*}
	In Fig.\ref{Fig_score}, we represent the learning result of the proposed model following the setting of $FixSegNum(\kappa = 5)$. The x-axis represents the estimation of $\boldsymbol{\mathcal{C}_{x_i}}$ , $\boldsymbol{\mathcal{C}_{z_i}}$ and $\boldsymbol{\mathcal{C}_{y_i|x_i}}$ respectively, while the y-axis shows the example count of each value. The threshold introduced in formula (\ref{threshold}) with mining ratio $r=\%10$ are drawn in the red dashed-lines. From the figure, it can be seen that from either $\boldsymbol{\mathcal{C}_{x_i}}$, $\boldsymbol{\mathcal{C}_{z_i}}$, or $\boldsymbol{\mathcal{C}_{y_i|x_i}}$ perspective, we can recover a long-tail distribution similar to the prediction error distribution in Fig.\ref{Fig_long}, with the majority samples are of higher likelihood estimation (normal or easy), while the minority samples are of relatively lower likelihood value (rare or hard). Under the circumstances, we can consider using the prediction errors from downstream models to validate whether the samples we have mined are critical.

	\subsection{Mining Results}
	The prediction error $Err$ and the percentage of error change $\triangle Err$ are listed in Tab.\ref{Tab_mining}. The downstream prediction model is set as RA-GAT $(RA)$, while the feature partition settings are fixed as $FixSegNum(\kappa = 5)$. From the table, it can be seen that for $D_X^*$, the prediction error grows 44.6\% when we mined 5\% samples from the original dataset. Similarly, from $D_Z^*$, we got 105.3\% of error change with 5\% mined samples. Combining the scores from $D_X^*$ and $D_Z^*$, our proposed hard set $D_{Y|X}^*$ reaches +108.1\% error compared to the average error on the original set. When the mining ratio increases to 20\%, we can still get a subset with +49\% error gain through $D_{Y|X}^*$. The prediction error on $D_{Y|X}^*$ is higher than $D_{rand}^r$ and $D_{ref}^r$, which indicates that our proposed mining method can find a more critical subset for downstream model $RA-GAT$.
\begin{figure*}[tb]
	\begin{center}
		\includegraphics[keepaspectratio=true,width=0.9\linewidth]{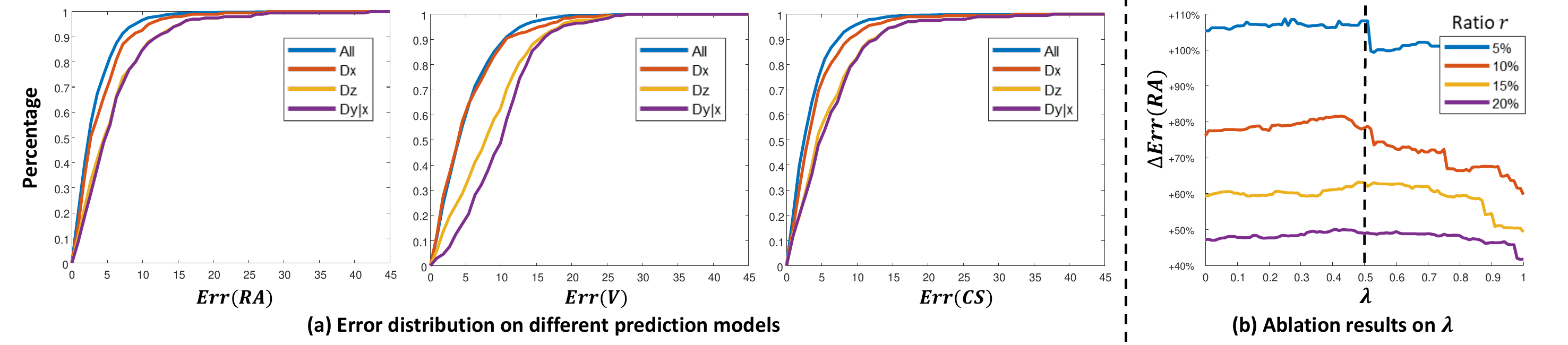}
	\end{center}
	\caption{(a) Prediction error coverage on different models with $r=10\%$. (b) Ablation results on different $\lambda$ settings. }
	\label{Fig_vis}
\end{figure*}	

	We then compare the error distribution on the mined subset on different models. The results are drawn in the formulation of coverage in Fig.\ref{Fig_vis}(a). It can be seen that our proposed $D_{Y|X}^*$ mining is of relatively higher prediction error on each model, the mining result is "critical" to any of the downstream models.
	  
	To better present such an advantage, we further list the $\triangle Err$ and mining coverage on different downstream models in Tab.\ref{Tab_modelfree}. The segmentation settings are fixed as $FixSegNum(\kappa = 5)$, and the mining results collected by $D_X^*$, $D_Z^*$ and $D_{Y|X}^*$ are both listed. From the table, it can be seen that on all models we got a obvious $\triangle Err$ when applying our mined samples compared to the original dataset. The $D_{Y|X}^*$ mining subset has similar $\triangle Err$ on all three models (about two times higher than average error when $r=5\%$). For mining coverage, our mined results have relatively close $Cov$ on all the three models, it means that the performance is not overfitting to any specific model, indicating the mining method as well as the mined samples are model-irrelevant - to some extent, unaffected by downstream prediction models. The $Cov^{ref}$ coverage describes the similarity between the mined samples with reference labels, when following $D_{Y|X}^*$ with $r=20\%$, the mined subset covers 59\% of the refence ground truth.

	\begin{table}[]
		\caption{$\triangle Err$ and $Cov$ on Different Feature Partition Settings with $r = 5\%$.}
		\centering
		\label{tab_ablation}
		\resizebox{1\linewidth}{!}
		{	\begin{tabular}{|c|ccc|ccc|}
				\hline
				Partition    & \multicolumn{3}{c|}{$FixSegNum$}                                      & \multicolumn{3}{c|}{$FixSegLen$}                                        \\ \hline
				$\kappa$        & \multicolumn{1}{c|}{$\kappa$ = 1}  & \multicolumn{1}{c|}{$\kappa$ = 3}  & $\kappa$ = 5   & \multicolumn{1}{c|}{$\kappa$ = 0.6} & \multicolumn{1}{c|}{$\kappa$ = 1.0} & $\kappa$ = 1.4 \\ \hline
				$\triangle Err (RA)$ & \multicolumn{1}{c|}{+81.5\%} & \multicolumn{1}{c|}{+99.3\%} & \textbf{+108.1\%} & \multicolumn{1}{c|}{+39.4\%}  & \multicolumn{1}{c|}{+69.6\%}  & +65.7\%  \\ \hline
				$Cov(RA)$     & \multicolumn{1}{c|}{26.0\%} & \multicolumn{1}{c|}{24.0\%} & \textbf{29.0\%}  & \multicolumn{1}{c|}{8.0\%}   & \multicolumn{1}{c|}{14.0\%}  & 16.0\%  \\ \hline
			\end{tabular}
	}
	\end{table}
	\subsection{Ablation Study}
	\label{exp_ablation}
	In this part, we conduct an ablation study to investigate the impact of different feature partition settings ($FixSegNum$ v.s. $FixSegLen$) and hyper-perameter $\lambda$ to our proposed mining method.

	\begin{figure*}[]
		\begin{center}
			\includegraphics[keepaspectratio=true,width=0.85\linewidth]{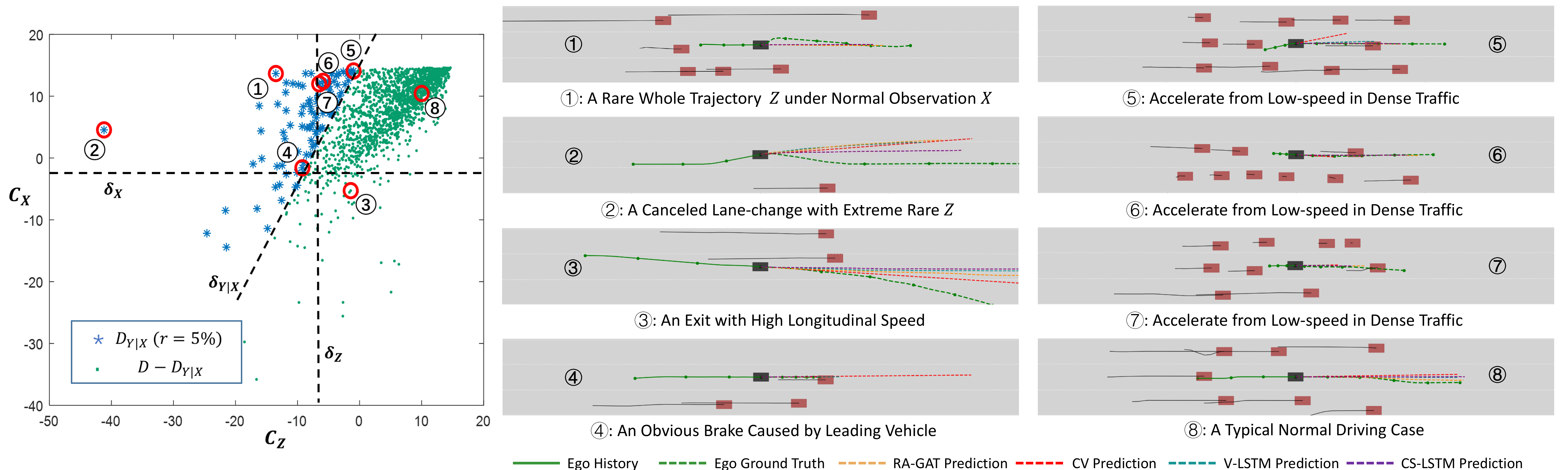}
		\end{center}
		\caption{Visualization of the $\boldsymbol{\mathcal{C}_{x_i}}$-$\boldsymbol{\mathcal{C}_{z_i}}$ space and the mined cases.}
		\label{fig_case}
	\end{figure*}
	\subsubsection{Hyper-parameter $\lambda$}
	In formula (\ref{lambda}), we introduce a hyper-parameter $\lambda$ to balance the influence of $\boldsymbol{\mathcal{C}_{z_i}}$ and $\boldsymbol{\mathcal{C}_{x_i}}$. To find the best $\lambda$ settings, we compare the prediction error change $\triangle Err$ on the range of [0-1] for $\lambda$. The results are presented in Fig.\ref{Fig_vis}(b). Note that when $\lambda = 0$, the $D_{Y|X}^*$ is the same as $D_{Z}^*$. From the figure it can be seen that when $\lambda > 0.5$ we got an evident performance drop on $\triangle Err$. We finally choose $\lambda = 0.5$ as the final settings for all the experimental results.

	\subsubsection{Feature Partition Setting}
	We present the $D_{Y|X}^*$ results with downstream model $RA-GAT$ for $r = 5\%$ on different feature partition settings in \ref{tab_ablation}. From the table, it can be seen that 1) The overall mining effectiveness of $FixSegNum$ is better than $FixSegLen$ from either error gain $\triangle Err$ or model coverage $Cov{(RA)}$ perspective. 2) The best segmentation setting in our scope is $FixSegNum(\kappa = 5)$, in which both $X$ and $Z$ are divided into 5 sub-sequences for feature extraction. It is notable that the segmentation method indicates a trade-off between over-smooth of features (lower sub-sequence number, longer sub-sequence length) and unstable features (shorter sub-sequence length, sensitive to noise in original data). We take $FixSegNum(\kappa = 5)$ as the final settings for all the experimental results.

	\subsection{Case Study}
	To better examine the effect of our mined examples, we conduct case studies in Fig.\ref{fig_case}. The left part shows the $D_{Y|X}^*$ result with $r = 5\%$. We marked out the threshold of $\delta_X$, $\delta_Z$ and $\delta_Y$ with black dashed lines for reference. The cases are marked by red circles in the coordinate system of $\boldsymbol{\mathcal{C}_{z_i}}-\boldsymbol{\mathcal{C}_{x_i}}$. 
	
	Case \textcircled{1} holds a relatively normal observation $X$, but has uncommon lateral deviation in $Z$. Our proposed method reports a relatively high $\boldsymbol{\mathcal{C}_{x_i}}$ (normal) with low $\boldsymbol{\mathcal{C}_{z_i}}$ (rare). Case \textcircled{2} reveals a canceled lane-change which is hard for models to predict. The example is captured by its extremely low $\boldsymbol{\mathcal{C}_{z_i}}$. Case \textcircled{3} shows a failure case when the ego vehicle exits the highway with high driving speed. Although $D_{Y|X}^*$ failed to capture such a case, we can still obtain the case via $D_X^*$ results. Case \textcircled{4} describe an obvious brake caused by a leading vehicle. In the feature space, the example is located near the threshold of $\delta_Y$. Case \textcircled{5}-\textcircled{7} are three similar scenario in which the ego vehicle accelerates from a dense traffic. These examples are projected into a nearby neighborhood in the space of $\boldsymbol{\mathcal{C}_{z_i}}-\boldsymbol{\mathcal{C}_{x_i}}$. Besides, these examples are not included in the result of $D_Z^*$, but our $D_{Y|X}^*$ results success in capturing the three examples. Finally, Case \textcircled{8} shows a normal driving case with both high $\boldsymbol{\mathcal{C}_{z_i}}$ and $\boldsymbol{\mathcal{C}_{z_i}}$.
	
	Through the case study, it can be seen that our proposed mining method can effectively dig out the uncommon driving examples from the raw dataset. These critical examples are relatively hard to predict for existing VTP models. We believe the discussion and further analysis on how to exploit these examples will bring more insights to vehicle trajectory prediction.  

	\section{Conclusion and Future works}
	This work proposes a model-irrelevant critical example mining method for vehicle trajectory prediction task. Through a unified feature extractor, the rareness of history observation and whole trajectory are estimated using flow-based generative models, based on which the hardness of the example for VTP task is evaluated. The samples with highest hardness are then collected as the mined results. Through experimental study, our main findings are:
	\begin{enumerate}
		\item The proposed method can effectively mine critical examples from the original dataset. When mining 5\% examples, the proposed hardness estimation returns a subset with +108\% error higher than average.
		\item The mining results of the proposed method are model-irrelevant. When applied to different models, the mined results of the proposed method have relatively higher error despite the change of downstream prediction model.
		\item The mining results are more likely to include uncommon cases such as sudden brake and canceled lane-change, which is indeed critical for both prediction error and scenario perspective.  
	\end{enumerate}
	
	The future studies of this work will involve validating the effect of the proposed mining method on a larger group of downstream models and on diverse vehicle trajectory prediction datasets. Besides, how to exploit the mined critical examples to improve the performance of prediction models will be further discussed.
	
	\bibliographystyle{IEEEtran}
	\bibliography{IEEEabrv,paper}

\begin{thebibliography}{10}
\providecommand{\url}[1]{#1}
\csname url@rmstyle\endcsname
\providecommand{\newblock}{\relax}
\providecommand{\bibinfo}[2]{#2}
\providecommand\BIBentrySTDinterwordspacing{\spaceskip=0pt\relax}
\providecommand\BIBentryALTinterwordstretchfactor{4}
\providecommand\BIBentryALTinterwordspacing{\spaceskip=\fontdimen2\font plus
\BIBentryALTinterwordstretchfactor\fontdimen3\font minus
  \fontdimen4\font\relax}
\providecommand\BIBforeignlanguage[2]{{%
\expandafter\ifx\csname l@#1\endcsname\relax
\typeout{** WARNING: IEEEtran.bst: No hyphenation pattern has been}%
\typeout{** loaded for the language `#1'. Using the pattern for}%
\typeout{** the default language instead.}%
\else
\language=\csname l@#1\endcsname
\fi
#2}}

\bibitem{ma2020artificial}
Y.~Ma, Z.~Wang, H.~Yang, and L.~Yang, ``Artificial intelligence applications in
  the development of autonomous vehicles: a survey,'' \emph{IEEE/CAA Journal of
  Automatica Sinica}, vol.~7, no.~2, pp. 315--329, 2020.

\bibitem{ding2023incorporating}
Z.~Ding and H.~Zhao, ``Incorporating driving knowledge in deep learning based
  vehicle trajectory prediction: A survey,'' \emph{IEEE Transactions on
  Intelligent Vehicles}, 2023.

\bibitem{mozaffari2020deep}
S.~Mozaffari, O.~Y. Al-Jarrah, M.~Dianati, P.~Jennings, and A.~Mouzakitis,
  ``Deep learning-based vehicle behavior prediction for autonomous driving
  applications: A review,'' \emph{IEEE Transactions on Intelligent
  Transportation Systems}, vol.~23, no.~1, pp. 33--47, 2020.

\bibitem{Brechtel2014}
S.~Brechtel, T.~Gindele, and R.~Dillmann, ``{Probabilistic decision-making
  under uncertainty for autonomous driving using continuous POMDPs},''
  \emph{2014 17th IEEE International Conference on Intelligent Transportation
  Systems, ITSC 2014}, pp. 392--399, 2014.

\bibitem{Bahram2016}
M.~Bahram, A.~Lawitzky, J.~Friedrichs, M.~Aeberhard, and D.~Wollherr, ``{A
  Game-Theoretic Approach to Replanning-Aware Interactive Scene Prediction and
  Planning},'' \emph{IEEE Transactions on Vehicular Technology}, vol.~65,
  no.~6, pp. 3981--3992, 2016.

\bibitem{zyner2019naturalistic}
A.~Zyner, S.~Worrall, and E.~Nebot, ``Naturalistic driver intention and path
  prediction using recurrent neural networks,'' \emph{IEEE transactions on
  intelligent transportation systems}, vol.~21, no.~4, pp. 1584--1594, 2019.

\bibitem{gu2021densetnt}
J.~Gu, C.~Sun, and H.~Zhao, ``Densetnt: End-to-end trajectory prediction from
  dense goal sets,'' in \emph{Proceedings of the IEEE/CVF International
  Conference on Computer Vision}, 2021, pp. 15\,303--15\,312.

\bibitem{ngiam2022scene}
J.~Ngiam, V.~Vasudevan, B.~Caine, Z.~Zhang, H.-T.~L. Chiang, J.~Ling,
  R.~Roelofs, A.~Bewley, C.~Liu, A.~Venugopal, \emph{et~al.}, ``Scene
  transformer: A unified architecture for predicting future trajectories of
  multiple agents,'' in \emph{International Conference on Learning
  Representations}, 2022.

\bibitem{deo2018convolutional}
N.~Deo and M.~M. Trivedi, ``Convolutional social pooling for vehicle trajectory
  prediction,'' in \emph{Proceedings of the IEEE Conference on Computer Vision
  and Pattern Recognition Workshops}, 2018, pp. 1468--1476.

\bibitem{ding2021ra}
Z.~Ding, Z.~Yao, and H.~Zhao, ``Ra-gat: Repulsion and attraction graph
  attention for trajectory prediction,'' in \emph{2021 IEEE International
  Intelligent Transportation Systems Conference (ITSC)}.\hskip 1em plus 0.5em
  minus 0.4em\relax IEEE, 2021, pp. 734--741.

\bibitem{zhang2023deep}
Y.~Zhang, B.~Kang, B.~Hooi, S.~Yan, and J.~Feng, ``Deep long-tailed learning: A
  survey,'' \emph{IEEE Transactions on Pattern Analysis and Machine
  Intelligence}, 2023.

\bibitem{he2009learning}
H.~He and E.~A. Garcia, ``Learning from imbalanced data,'' \emph{IEEE
  Transactions on knowledge and data engineering}, vol.~21, no.~9, pp.
  1263--1284, 2009.

\bibitem{zhou2022long}
W.~Zhou, Z.~Cao, Y.~Xu, N.~Deng, X.~Liu, K.~Jiang, and D.~Yang, ``Long-tail
  prediction uncertainty aware trajectory planning for self-driving vehicles,''
  in \emph{2022 IEEE 25th International Conference on Intelligent
  Transportation Systems (ITSC)}.\hskip 1em plus 0.5em minus 0.4em\relax IEEE,
  2022, pp. 1275--1282.

\bibitem{makansi2021exposing}
O.~Makansi, {\"O}.~Cicek, Y.~Marrakchi, and T.~Brox, ``On exposing the
  challenging long tail in future prediction of traffic actors,'' in
  \emph{Proceedings of the IEEE/CVF International Conference on Computer
  Vision}, 2021, pp. 13\,147--13\,157.

\bibitem{kozerawski2022taming}
J.~Kozerawski, M.~Sharan, and R.~Yu, ``Taming the long tail of deep
  probabilistic forecasting,'' \emph{arXiv preprint arXiv:2202.13418}, 2022.

\bibitem{jiang2022improving}
C.~M. Jiang, M.~Najibi, C.~R. Qi, Y.~Zhou, and D.~Anguelov, ``Improving the
  intra-class long-tail in 3d detection via rare example mining,'' in
  \emph{European Conference on Computer Vision}.\hskip 1em plus 0.5em minus
  0.4em\relax Springer, 2022, pp. 158--175.

\bibitem{gao2020vectornet}
J.~Gao, C.~Sun, H.~Zhao, Y.~Shen, D.~Anguelov, C.~Li, and C.~Schmid,
  ``Vectornet: Encoding hd maps and agent dynamics from vectorized
  representation,'' in \emph{Proceedings of the IEEE/CVF Conference on Computer
  Vision and Pattern Recognition}, 2020, pp. 11\,525--11\,533.

\bibitem{lee2019joint}
D.~Lee, Y.~Gu, J.~Hoang, and M.~Marchetti-Bowick, ``Joint interaction and
  trajectory prediction for autonomous driving using graph neural networks,''
  \emph{arXiv preprint arXiv:1912.07882}, 2019.

\bibitem{altche2017lstm}
F.~Altch{\'e} and A.~de~La~Fortelle, ``An lstm network for highway trajectory
  prediction,'' in \emph{2017 IEEE 20th international conference on intelligent
  transportation systems (ITSC)}.\hskip 1em plus 0.5em minus 0.4em\relax IEEE,
  2017, pp. 353--359.

\bibitem{messaoud2019relational}
K.~Messaoud, I.~Yahiaoui, A.~Verroust-Blondet, and F.~Nashashibi, ``Relational
  recurrent neural networks for vehicle trajectory prediction,'' in \emph{2019
  IEEE Intelligent Transportation Systems Conference (ITSC)}.\hskip 1em plus
  0.5em minus 0.4em\relax IEEE, 2019, pp. 1813--1818.

\bibitem{xu2022group}
D.~Xu, X.~Shang, Y.~Liu, H.~Peng, and H.~Li, ``Group vehicle trajectory
  prediction with global spatio-temporal graph,'' \emph{IEEE Transactions on
  Intelligent Vehicles}, 2022.

\bibitem{huang2022multi}
Z.~Huang, X.~Mo, and C.~Lv, ``Multi-modal motion prediction with
  transformer-based neural network for autonomous driving,'' in \emph{2022
  International Conference on Robotics and Automation (ICRA)}.\hskip 1em plus
  0.5em minus 0.4em\relax IEEE, 2022, pp. 2605--2611.

\bibitem{yu2023dataset}
R.~Yu, S.~Liu, and X.~Wang, ``Dataset distillation: A comprehensive review,''
  \emph{arXiv preprint arXiv:2301.07014}, 2023.

\bibitem{wang2018dataset}
T.~Wang, J.-Y. Zhu, A.~Torralba, and A.~A. Efros, ``Dataset distillation,''
  \emph{arXiv preprint arXiv:1811.10959}, 2018.

\bibitem{harshvardhan2020comprehensive}
G.~Harshvardhan, M.~K. Gourisaria, M.~Pandey, and S.~S. Rautaray, ``A
  comprehensive survey and analysis of generative models in machine learning,''
  \emph{Computer Science Review}, vol.~38, p. 100285, 2020.

\bibitem{kingma2019introduction}
D.~P. Kingma, M.~Welling, \emph{et~al.}, ``An introduction to variational
  autoencoders,'' \emph{Foundations and Trends{\textregistered} in Machine
  Learning}, vol.~12, no.~4, pp. 307--392, 2019.

\bibitem{goodfellow2020generative}
I.~Goodfellow, J.~Pouget-Abadie, M.~Mirza, B.~Xu, D.~Warde-Farley, S.~Ozair,
  A.~Courville, and Y.~Bengio, ``Generative adversarial networks,''
  \emph{Communications of the ACM}, vol.~63, no.~11, pp. 139--144, 2020.

\bibitem{rezende2015variational}
D.~Rezende and S.~Mohamed, ``Variational inference with normalizing flows,'' in
  \emph{International conference on machine learning}.\hskip 1em plus 0.5em
  minus 0.4em\relax PMLR, 2015, pp. 1530--1538.

\bibitem{dinh2014nice}
L.~Dinh, D.~Krueger, and Y.~Bengio, ``Nice: Non-linear independent components
  estimation,'' \emph{arXiv preprint arXiv:1410.8516}, 2014.

\bibitem{I80}
J.~Colyar and J.~Halkias, ``Us highway 80 dataset,'' Federal Highway
  Administration (FHWA), Tech. Rep., 2007.

\bibitem{US101}
------, ``Us highway 101 dataset,'' Federal Highway Administration (FHWA),
  Tech. Rep., 2007.

\end{thebibliography}
\end{document}